\documentclass[11pt,a4paper]{article}
\usepackage{authblk}
\usepackage[hyperref]{acl2021}
\usepackage{times}
\usepackage{latexsym}
\usepackage{subfigure}
\usepackage{amsmath}
\usepackage{multirow}
\usepackage{lscape} 
\usepackage{graphicx}
\usepackage{multirow}
\usepackage{amsfonts}

\renewcommand{\UrlFont}{\ttfamily\small}

\usepackage{microtype}

\aclfinalcopy % Uncomment this line for the final submission
\def\aclpaperid{1571} %  Enter the acl Paper ID here

\title{RetroGAN: A Cyclic Post-Specialization System for Improving Out-of-Knowledge and Rare Word Representations}

\author[1]{\textbf{Pedro Colon-Hernandez}}
\affil[1]{MIT Media Lab\authorcr\textit{\{pe25171,cynthiab\}@media.mit.edu}}
\author[2]{\textbf{Yida Xin}}
\affil[2]{Boston University\authorcr\textit{\{yxin,spchin\}@cs.bu.edu}}
\author[3]{\textbf{Henry Lieberman}}
\affil[3]{MIT CSAIL\authorcr \textit{lieber@media.mit.edu}}
\author[4]{\\\textbf{Catherine Havasi}}
\affil[4]{Basis Technologies\thanks{\hspace{0.15cm}Work done while at the MIT Media Lab}\authorcr \textit{havasi@basistech.com}}
\author[1]{\textbf{Cynthia Breazeal}}
\author[2]{\textbf{Peter Chin}}

\date{}

\begin{document}
\maketitle
\begin{abstract}

%% 1. What's the problem?
%% 2. Why is it important?
%% 3. What's the SOTA?
%% 4. What's my big idea of solving it?
%% 5. What did I do to test it?
%% 6. What are the results and takeaways?

%Word embeddings (e.g. FastText) can provide a wide coverage of concepts, however their reliance on words appearing in similar contexts makes the knowledge relatively weak evidence for inference. On the other hand, human-curated knowledge bases (e.g. WordNet and ConceptNet) provide stronger explicit evidence of relationships between concepts, however they have relatively lower coverage of concepts. 
Retrofitting is a technique used to move word vectors closer together or further apart in their space to reflect their relationships in a Knowledge Base (KB). However, retrofitting only works on concepts that are present in that KB.  RetroGAN uses a pair of Generative Adversarial Networks (GANs) to learn a one-to-one mapping between concepts and their retrofitted counterparts. It applies that mapping (post-specializes) to handle concepts that do not appear in the original KB in a manner similar to how some natural language systems handle out-of-vocabulary entries.
% We present a system called RetroGAN, which learns a guaranteed one-to-one mapping from original to retrofitted embeddings which permits generalizing retrofitted knowledge (post-specialization). RetroGAN uses a pair of GANs to post-specialize out-of-KB (OOK) concepts in a manner similar to how some natural language systems handle out-of-vocabulary (OOV) entries.
We test our system on three word-similarity benchmarks and a downstream sentence simplification task, and achieve the state of the art (CARD-660). Altogether, our results demonstrate our system's effectiveness for out-of-knowledge and rare word generalization.
\end{abstract}

\section{Introduction}
\label{Introduction}

Retrofitting word embeddings with a KB \cite{faruqui2014retrofitting,speer2016ensemble, mrkvsic2017semantic} means taking a vector space of word embeddings and finding a mapping that moves some of these word vectors closer together and others further apart, such that these vectors' new positions in the vector space are in better agreement with the relationships between the same words (a.k.a., concepts) in a KB \cite{speer2016ensemble,mrkvsic2017semantic}. However, the retrofitting process can only work on concepts that are actually present in the KB (a.k.a., constraints), which means that retrofitting can get us improved performance in semantic tasks only on the overlapping vocabulary between the KB and the word embeddings. Post-specialization\cite{vulic2018post,kamath2019specializing} is a solution to this problem; it is a series of techniques that try to (1) learn the mapping that retrofitting establishes and (2) generalize the mapping to the rest of the embedding vocabulary. 

We develop and present a post-specialization system called RetroGAN that builds upon the approach presented as AuxGAN \cite{ponti2018adversarial} by extending it to have a pair of Generative Adversarial Networks (GANs) \cite{goodfellow2014generative}. A regular GAN minimizes the loss when learning the function for post-specialization. Our pair works in a cyclic manner to minimize the losses of both the post-specialization and the inverse to ensure that there is a one-to-one mapping between the two domains. This constrains the outputs for unseen data in both domains and leads to achieving higher performance for unseen concepts. 
%The contribution that we present in this paper is RetroGAN: a CycleGAN-like post-specialization system that improves the ability of word embedding approaches to incorporate semantic constraints. 
% We utilize RetroGAN to learn an Attract-Repel mapping (\cite{mrkvsic2017semantic} to post-specialize a corpus of Fasttext embeddings.   However, we still run into the problem of being able to get the relationships between concepts back in an explicit manner.  By retrofitting and then post-specializing, some semantic information from the knowledge base and/or constraints is being fused with the one found in the embeddings, but it is stored as a vector for the concept rather than as a graph.  
\begin{table*}[]
    \small
    \centering
    \label{retrogan_testing}
    \centering
    \begin{tabular}{ |p{1cm}|p{6.3cm}|p{6.3cm}|  }
        \hline
        Word & Distributional Neighbor & Retrofitted Neighbor\\
        \hline
        Dog &dog, dogs, puppy, pup , canine, pet, doggie, beagle, dachshund, cat & dog, beagle, pooch, dachshund, puppy, mutt, poodle, Rottweiler, canine, labrador\\
        \hline
        Doggo & pooch, doggies, bae, chihuahua, rad, pug, kitty, dane, furbabies, \texttt{\symbol{92}uf602} & doggies, pooch, dachshund, four-legged, Yorkie, corgi, whippet, amigos, Weimaraner, Dog\\
        \hline
    \end{tabular}
    \caption{Results of the 10 most similar embeddings for “dog” and “doggo” for FastText embeddings. The distributional neighbors are the closest embeddings in the original distributional space and the retrofitted neighbors are the closest in the RetroGAN post-specialized space.  We can see that ``doggo'' was near slangs such as ``bae'' and ``furbabies'', but after post-specialization, it gets closer to words that we regard as semantically similar to ``dog.'' The one-to-one mapping that RetroGAN provides is key to being able to incorporate useful semantic information into rare-words possibly like ``doggo".}
    \label{table:dogneighbors}
\end{table*}

\section{Related Work}
Within the field of retrofitting, work has been done in exploring the various ways of infusing constraints or KBs into word embeddings.  The original work by \cite{faruqui2014retrofitting} only used synonymy relationships but not antonymy relationships, which meant that word embeddings with similar (synonymous) semantics in the KB would be pulled together, but word embeddings with dissimilar (antonymous) semantics would not be separated. The Attract-Repel work by \cite{mrkvsic2017semantic} addressed this shortcoming by incorporating antonymy relationships in a retrofitting procedure: synonymous embeddings are \emph{attracted} to each other, while antonymous embeddings are \emph{repelled} against each other.  This line of work was continued with the work done by Lexical Entailment Attract-Repel \cite{vulic2017specialising}(LEAR), which looks to add the asymmetric lexical entailment relationship to Attract-Repel.
\footnote{We do not use LEAR because in the original work, it did not alter the similarity tasks results, but they can be exchanged. 
% However we can replace Attract-Repel for LEAR.
} 

Building on these works, a series of techniques called \emph{post-specialization} were developed.  These techniques consist on utilizing neural models to learn retrofitting mappings such as \cite{glavavs2018explicit} and \cite{ponti2018adversarial,kamath2019specializing} which use a Deep Feed-forward Neural Network and a Generative Adversarial Network (GAN) respectively.  Post-specialization permits, provided a static word embedding, to generate its retrofitted counterpart on the fly with a trained system. A concrete example is in table \ref{table:dogneighbors}.  

As it stands, attention has been shifted to using contextual embeddings such as those produced from BERT \cite{devlin2018bert} on downstream tasks.  Only recently have there been efforts in incorporating external, KB assertions into pre-trained transformer-based systems (e.g., KnowBERT \cite{peters2019knowledge},  Align-mask-select \cite{ye2019align}, and LIBERT \cite{lauscher2019informing}). LIBERT bridges contextual and retrofitted embeddings by leveraging the knowledge in retrofitted embeddings to find lexical tuples that are fed into BERT to focus on their lexical information.

GANs have been utilized extensively in the image domain to create lifelike images.  CycleGAN \cite{zhu2017unpaired} and other cyclic systems \cite{kim2017learning} have been utilized to perform style transfer (i.e. apply certain distinctive characteristics from one image domain into another).  CycleGAN serves to learn a, possibly unpaired, one-to-one mapping from one domain to another.  To effectively utilize paired data, the work by \cite{tripathy2018learning} modifies the CycleGAN architecture to include a conditional cyclic loss in which new discriminators are conditioned to determine if a generated sample is real or not based on a given, possibly paired, input.  This in turn permits leveraging paired data to improve the one-to-one mapping.

\section{RetroGAN}
% Retrofitting is a process in which word embeddings are post-processed (a.k.a., specialized) and optimized to favor certain optimization criteria. In one of the most recent works on retrofitting\cite{mrkvsic2017semantic}, the criteria enforce that (1) concepts that are connected through an attract (e.g., synonymy) relationship should have similar (i.e., closer together) embeddings, and (2) concepts that are connected through a repel (e.g., antonymy) relationship should have dissimilar (i.e., further apart) embeddings. Retrofitting greatly improves the performance of static embeddings in word similarity and the performance of downstream tasks such as lexical simplification. However, the retrofitting operation can only be performed on embeddings whose concepts are explicitly present in a KB.  This typically leads to having high performance on a small subset of the entire vocabulary of a word embedding corpus. (For example, for the complete Glove-CC and Fasttext-CC this subset is only 2.65\% and 2.99\% respectively.)

% The line of work by \cite{glavavs2018explicit, ponti2018adversarial,kamath2019specializing} presents what is called \emph{post-specialization}: a technique that uses neural architectures that learn to apply a retrofitting mapping for word embeddings that are unseen in the retrofitting constraints. Intuitively, post-specialization expands the constraints or KB that are used to retrofit because it tries to make similar embeddings have similar semantically retrofitted counterparts. 

RetroGAN is a system that builds on \cite{ponti2018adversarial} by utilizing a CycleGAN-like architecture (i.e., we use a pair of GANs cyclicly but our layers are different from the original CycleGAN). We chose the CycleGAN-like architecture because, in our domain, the cycle-consistency constraints can enforce a one-to-one mapping from original embeddings to retrofitted embeddings. This mapping guarantees that unseen concepts will have their own, unique retrofitted counterparts. We use RetroGAN to learn the mapping of Attract-Repel \cite{mrkvsic2017semantic} retrofitting (with the \emph{synonymy} and \emph{antonymy} constraints from the Attract-Repel paper\cite{mrkvsic2017semantic}) on a subset of static word embeddings (i.e.., FastText \cite{bojanowski2017enriching},  and Numberbatch \cite{speer2017conceptnet}), and perform post-specialization on the entire set.  
% We focus particularly on static embeddings  rather than contextual embeddings (e.g., BERT \cite{devlin2018bert} and other Transformer-based systems), because there is still no decisive way of adding KB constraints into these systems to enforce explicit semantics. 
% We note, however, that there are efforts in this area that seem promising with regards to injecting constraints \cite{wang2020k,peters2019knowledge,lauscher2019informing}.  
% \begin{figure}[]
%   \centering
%   \includegraphics[width=0.5\textwidth,]{BlankDiagram-2.png}
%   \caption{RetroGAN System Architecture}
%   \label{fig:rg_arch}
% \end{figure}
\subsection{Model \& Architecture}

RetroGAN consists of two GANs that interplay to balance a combination of losses to transform a particular word embedding $x_i \in X$ from its original domain $X$ to its counterpart $y_i \in Y$ in the retrofitted domain $Y$, and vice-versa.  In both GANs that we employ, the generator consists of an input layer followed by 2 hidden dense layers with 2048 neurons and each followed by a dropout layer (with a percentage of 0.2 for the dropouts), and a final linear output layer with the same dimensionality as the input. The output of this layer, for the trained $G:X	\rightarrow
Y$ produces the post-specialized embeddings (i.e. a batch of 32 FastText embeddings produces 32 post-specialized embeddings). The hidden layers employ the ReLU \cite{nair2010rectified} activation function. 
Our discriminators have a similar structure (an input layer, 2 hidden layers with dropout but a percentage of 0.3), however, the second hidden layer is followed by a batch normalization layer and the output is a single neuron with a sigmoid activation.  The reason for the batch normalization layer was to stabilize the training. We also utilized a third and fourth conditional discriminator following \cite{tripathy2018learning}, to leverage the cyclic architecture on paired data.

% \begin{table*}[]
% \small
% \centering
% \begin{tabular}{|c|c|c|c|c|c|}

% \hline
%                                         %  \\ \hline
% Model &
%   Epochs &
%   Epochs &
%   \begin{tabular}[c]{@{}c@{}}Distributional \\ (FastText)\end{tabular} &
%   \begin{tabular}[c]{@{}c@{}}Attract-Repel\\ (FastText)\end{tabular} &
%   \begin{tabular}[c]{@{}c@{}}Numberbatch+ Attract-Repel\\ (FastText)\end{tabular} \\ \hline
% AuxGAN   & 1(1M Iterations)     & ADAM(lr=0.001) & 0.42295 & 0.2691           & 0.15671         \\ \hline
% AuxGAN   & 1(1M Iterations)     & SGD(lr=0.1)    & 0.42295 & \textbf{0.45231} & 0.46184         \\ \hline
% AuxGAN   & 10(10M Iterations)   & ADAM(lr=0.001) & 0.42295 & 0.20228          & 0.3785          \\ \hline
% AuxGAN   & 10(10M Iterations)   & SGD(lr=0.1)    & 0.42295 & 0.29351          & 0.44376         \\ \hline
% RetroGAN & 150($\sim$300k Iterations) & ADAM           & 0.42295 & 0.36508          & \textbf{0.5497} \\ \hline
% \end{tabular}
% \label{table:card660}
% \caption{Cambridge Rare word evaluation\cite{pilehvar2018card} results}
% \end{table*}

% Please add the following required packages to your document preamble:
% Please add the following required packages to your document preamble:
% \usepackage{multirow}

\begin{table*}[]
\small
\centering
\begin{tabular}{l|r|r|r|r|r|r|r|r|r|r|}
\cline{2-11}
                                     & \multicolumn{4}{l|}{Disjoint}                                                                         & \multicolumn{6}{l|}{Full}                                                                                                                                     \\ \cline{2-11} 
                                     & \multicolumn{2}{l|}{FT-CC, A-R}                   & \multicolumn{2}{l|}{FT-CC, A-R+NB}                & \multicolumn{3}{l|}{FT-CC, Attract-Repel}                                     & \multicolumn{3}{l|}{FT-CC, A-R+NB}                                            \\ \hline
\multicolumn{1}{|l|}{Models}         & \multicolumn{1}{l|}{SL} & \multicolumn{1}{l|}{SV} & \multicolumn{1}{l|}{SL} & \multicolumn{1}{l|}{SV} & \multicolumn{1}{l|}{SL} & \multicolumn{1}{l|}{SV} & \multicolumn{1}{l|}{C660} & \multicolumn{1}{l|}{SL} & \multicolumn{1}{l|}{SV} & \multicolumn{1}{l|}{C660} \\ \hline
\multicolumn{1}{|l|}{Distributional} & 0.4644                  & 0.3649                  & 0.4499                  & 0.3643                  & 0.4644                  & 0.3649                  & 0.2973                    & 0.4499                  & 0.3643                  & 0.1068                    \\ \hline
\multicolumn{1}{|l|}{Attract-Repel}  & 0.4644                  & 0.3649                  & 0.4499                  & 0.3643                  & \textbf{0.7790}         & \textbf{0.7632}         & 0.3768                    & 0.7748                  & \textbf{0.7667}         & 0.2203                    \\ \hline
\multicolumn{1}{|l|}{AuxGAN}         & \textbf{0.6127}         & 0.4641                  & 0.6116                  & 0.5331                  & 0.6901                  & 0.5756                  & 0.3899                    & 0.6565                  & 0.5872                  & 0.2088                    \\ \hline
\multicolumn{1}{|l|}{RetroGAN}       & 0.6028                  & \textbf{0.4702}         & \textbf{0.6648}         & \textbf{0.5971}         & 0.7717                  & 0.7192                  & \textbf{0.5240}           & \textbf{0.7960}         & 0.7483                  & \textbf{0.5581}           \\ \hline
\end{tabular}

\caption{\textit{Word similarity tests results}: We run two distinct scenarios in which the words present in SL and SV are present (Full) in Attract-Repel (A-R) retrofitting constraints or not (Disjoint). The Distributional row represents the results of the tests using the publicly available embeddings (FT-CC). The results are the Spearman correlation ($\rho$) between the Cosine distance of the embeddings we are evaluating and the human similarity measurements.  The results for AuxGAN and RetroGAN are the average of 3 runs.}
\label{table:results}

\end{table*}

A novelty in RetroGAN is the combination of cyclic and non-cyclic optimization objectives: the regular adversarial loss for both GANs ($L_{GAN}$); the cyclic loss for both generators ($L_{CYC}$); the identity loss for both generators ($L_{ID}$); the max margin loss similar to \cite{weston2011wsabie,ponti2018adversarial} for both the generators and additionally for the cycle of generators ($L_{MM}$); and the conditional cycle consistency loss ($L_{cCYC}$) introduced in \cite{tripathy2018learning}.  The combined objective has the following form:  
\begin{equation}
\small
\begin{split}
L(G, F, D_X, D_Y ) =L_{GAN}(G, D_Y , X, Y ) +\\ L_{GAN}(F, D_X, X, Y) + \lambda L_{CYC}(G, F)+\\ \gamma L_{ID}(G, F, X, Y)+  L_{MM}(G,F,X,Y)+\\ \varsigma L_{cCYC}(G,F,D_{cX},D_{cY},X,Y)
\end{split}
\end{equation}
where $G:X	\rightarrow Y$ is the generator that maps the source domain $X$ of plain word embeddings to the target domain $Y$ of retrofitted word embeddings; $F:Y \rightarrow X$ is the generator that does the opposite; $D_X$ and $D_Y$ are the discriminators for the corresponding domains; and $D_{cX},D_{cY}$ are our cycle conditional discriminators.  For brevity, we only go into details on $L_{MM}$ and $L_{cCYC}$. The other losses are the standard ones found in their respective works: $L_{GAN}$ is the adversarial loss from \cite{goodfellow2014generative}.
% \begin{equation}
% \small
% \begin{split}
%     L_{GAN}(G,D_Y,X,Y) = \mathbb{E}_{y\sim p_{data}(y)})[log D_Y(y)]+\\ \mathbb{E}_{x\sim p_{data}(x)})[log(1-D_Y (G(x))],
% \end{split}
% \end{equation}
% where $p_{data}(x)$ and $p_{data}(y)$ are the data distributions.
$L_{CYC}$ is the cycle consistency loss from \cite{zhu2017unpaired}
% \begin{equation}
% \small
% \begin{split}
% L_{CYC}(G, F) = \mathbb{E}_{x\sim p_{data}(x)}[||F(G(x))- x||_1]
% +\\ \mathbb{E}_{y\sim p_{data}(y)}
% [||G(F(y))- y||_1],
% \end{split}
% \end{equation}
with a scaling factor of $\lambda$ (which we set to 1); and $L_{ID}$ is the identity loss from \cite{zhu2017unpaired},
% \begin{equation}
% \small
% \begin{split}
% L_{ID}(G, F)=\mathbb{E}_{y\sim p_{data}(y)}[||G(y)- y||_1] +\\\mathbb{E}_{x\sim p_{data}(x)}[||F(x)-x||_1],
% \end{split}
% \end{equation}
which we scale with $\gamma$ (which we set to 0.01). $L_{ID}$ serves as a check of whether the embedding is already in the correct domain. $L_{MM}$ is the max margin loss with random confounders as used by \cite{ponti2018adversarial}, and as a novel aspect, we add a cyclic margin loss:
\begin{equation}
\small
\begin{split}
L_{MM}(G,F,X,Y) = 
\Sigma^{||x||}_{i=1}
\Sigma^k_{j=1|j\neq i}
\tau [\\(\delta_{MM}-cos(G(x_i),y_i)+cos(G(x_i), y_j))+\\
(\delta_{MM}-cos(F(y_i),x_i)+cos(F(y_i), x_j))+\\
\pmb{(\delta_{MM}-cos(G(F(y_i)),y_i)+cos(G(F(y_i)), y_j))}+\\
\pmb{(\delta_{MM}-cos(F(G(x_i)),x_i)+cos(F(G(x_i)), x_j))}]
\end{split}
\label{eq:mmloss}
\end{equation}
Equation \ref{eq:mmloss}, 
% is composed of 4 elements, which have the same structure.  Within that structure, in the first term we try to increase the cosine similarity for generated embeddings up to a margin $\delta_{MM}$ (which we set to 1); in the second term we make the generated embeddings distance themselves from other embeddings (random confounders), and this is done for an entire batch of $k$ (which we set to 25) confounders. The sum of the 4 elements is scaled by $\tau$ (which we set to 1). I
intuitively, tries to make generated embeddings similar to their gold-standard and different from confounders.  RetroGAN further enforces this constraint across the cycle.
%of $X\rightarrow Y\rightarrow X$ and vice versa.  
Lastly, we have $L_{cCYC}$ which is the conditional cycle loss \cite{tripathy2018learning}\footnote{In future work we will additionally incorporate the paired conditional adversarial loss.}, which we scale with $\varsigma$ (set to 1):
\begin{equation}
\small
\begin{split}
    L_{cCYC}(G,F,D_{cX},D_{cY},X,Y) =\\ \mathbb{E}_{x\sim p_{data}} [log(D_{cX}(G(x),x))]+\\\mathbb{E}_{x\sim p_{data}} [log(1-D_{cX}(G(x),F(G(x))))]+\\
    \mathbb{E}_{y\sim p_{data}} [log(D_{cY}(F(y),y))]+\\\mathbb{E}_{x\sim p_{data}} [log(1-D_{cY}(F(y),G(F(y))))]
\end{split}
\end{equation}
\subsection{Experimental Setup}
To train our system we utilize the ADAM \cite{kingma2014adam} optimizer with a learning rate of 5e-5 for the generators and 1e-4 for the non-conditional discriminators. We do not train the discriminators used in the regular GAN loss, and instead train the ones in the conditional cycle consistency loss. We also note that we did not perform explicit fine tuning of the scaling parameters, but we will do so in future work through a grid search. We train for 312,500 mini-batches which is the equivalent to the AuxGAN training, using a batch size of 32.

% \begin{figure}[h]
    
%   \centering
%   \includegraphics[width=1\linewidth]{Resultstable.png}
%   \renewcommand{\figure}{\caption}{Table 2: SimLex and SimVerb evaluation on word embeddings}
%   \label{fig:tablebig}
% \end{figure}

% We now describe the embeddings that we use in our tests. 
% We utilized the Glove embeddings trained on the Common Crawl with 840 billion tokens and dimensionality of 300 (Glove-CC in our tests).  
In our tests we use the English Common Crawl FastText with sub-word information (FT-CC) and Numberbatch 19.08 (NB) to see how performance would be affected by using embeddings that were already retrofitted with a large KB. We ran the Attract-Repel \cite{mrkvsic2017semantic}\footnote{We use the default settings found in \linebreak \url{https://github.com/nmrksic/attract-repel}} procedure on all these embeddings then proceeded to perform our post-specialization tests on learning the mapping from FT-CC to the resulting retrofitted embeddings.   

\begin{table*}[]
\small
\centering
\begin{tabular}{l|r|r|r|r|r|r|r|r|r|r|r|r|}
\cline{2-13}
                                    & \multicolumn{3}{r|}{5\%}                                                      & \multicolumn{3}{r|}{10\%}                                                     & \multicolumn{3}{r|}{25\%}                                                     & \multicolumn{3}{r|}{50\%}                                                     \\ \cline{2-13} 
                                    & \multicolumn{1}{l|}{SL} & \multicolumn{1}{l|}{SV} & \multicolumn{1}{l|}{C660} & \multicolumn{1}{l|}{SL} & \multicolumn{1}{l|}{SV} & \multicolumn{1}{l|}{C660} & \multicolumn{1}{l|}{SL} & \multicolumn{1}{l|}{SV} & \multicolumn{1}{l|}{C660} & \multicolumn{1}{l|}{SL} & \multicolumn{1}{l|}{SV} & \multicolumn{1}{l|}{C660} \\ \hline
\multicolumn{1}{|l|}{Attract-Repel} & 0.347                   & 0.355                   & 0.113                     & 0.550                   & 0.589                   & 0.187                     & 0.701                   & 0.700                   & 0.217                     & \textbf{0.759}          & \textbf{0.747}          & 0.252                     \\ \hline
\multicolumn{1}{|l|}{AuxGAN}        & 0.615                   & 0.510                   & 0.453                     & 0.667                   & 0.569                   & 0.470                     & 0.679                   & 0.581                   & 0.475                     & 0.685                   & 0.600                   & 0.490                     \\ \hline
\multicolumn{1}{|l|}{RetroGAN}      & \textbf{0.624}          & \textbf{0.538}          & \textbf{0.489}            & \textbf{0.701}          & \textbf{0.652}          & \textbf{0.493}            & \textbf{0.738}          & \textbf{0.690}          & \textbf{0.502}            & 0.755                   & 0.716                   & \textbf{0.511}            \\ \hline
\end{tabular}
\caption{Out of knowledge tests: We evaluate the performance of retrofitting and post specialization at varying percentages of test words seen in constraints. The full table can be seen in Appendix \ref{appendix:fulloovtest}}
\label{table:ooktests}
\end{table*}
We ran the word similarity benchmarks: SimLex (SL)\cite{hill2015simlex} SimVerb (SV)\cite{gerz2016simverb}, and the Cambridge Rare Word (C660) dataset \cite{pilehvar2018card}. We utilize the Disjoint (evaluating words which were \textbf{not seen} in the constraints) and Full (evaluating words which were \textbf{seen} in constraints) settings from \cite{ponti2018adversarial} for SL and SV, and evaluate C660 on the Full setting to test performance on rare words. The maximum values for the similarity benchmarks while training are listed in table \ref{table:results}.

We trained the publicly available AuxGAN model on 10 epochs of 1M iterations (which in AuxGAN is a single embedding pair rather than a batch of pairs) with both plain stochastic gradient descent (SGD) and ADAM (learning rate of 0.1) and selected the best performing one (ADAM) to compensate for convergence speed discrepancies. Additionally, similar to \cite{ponti2018adversarial} we also evaluated on Light-LS \cite{glavavs2015simplifying} with the default dataset \cite{horn2014learning} to test downstream performance. 
We evaluate the accuracy of simplification substitutions(i.e., the amount of words that are substituted correctly when compared to a gold standard). We utilize the first 500k words in FT-CC and the complete Numberbatch (generating the vocabulary's FastText embeddings as the distributional model). To test the retrofitted embeddings we substitute them in the original set. 
\subsection{Results \& Discussion}
 RetroGAN outperforms AuxGAN in the majority of similarity benchmarks.  We note that RetroGAN sets the state of the art on the rare-words benchmark (C660)(previously, to the best of our knowledge, it was 0.543 and 0.55 in \cite{yang2019out, NobukazuFukuda2020Based}). 
In the similarity results for Full , we note the same observations that were noted in AuxGAN: there are some inconsistent gains and losses, which may be due to the combination of loss functions which may make the systems imprecise; although they spread the knowledge throughout the embeddings, they lose some precision when compared with the original retrofitted embeddings.  The results for the lexical simplification (Light-LS) can be seen in table \ref{tab:lightlstest} where RetroGAN dominates.
 
We wanted to compare the out-of-knowledge (OOK) performance more in depth and to do this, we joined the words in SimLex (SL) and SimVerb (SV) and selected increasingly larger amounts of them (\{5,10,25,50,75,100\}\%). We then selected the constraints that included these words, trained RetroGAN and AuxGAN with these constraints, and evaluated performance on SL, SV, and C660.  Part of this can be seen in table \ref{table:ooktests}. We see that RetroGAN's performance increases every time that new constraints are added, whereas AuxGAN's performance begins to peak after 25\% of the constraints which may indicate more efficient knowledge distribution thanks to the cyclic system. Later on, the performance of RetroGAN kept increasing, but was less than the base retrofitted embeddings, possibly because of the lack of precision from the combination of losses. Lastly, we performed a small ablation study (Appendix \ref{appendix:ablationtests}) on RetroGAN's losses. We note that the max-margin loss from \cite{ponti2018adversarial} is necessary for high performance in all the tests.  We also notice that the cyclic (cyclic max-margin and cycle conditional discriminator) losses are essential for improved performance on the OOK and rare-word similarity benchmarks. We also see that the removal of the cyclic max-margin loss speeds up early learning and its addition stabilizes later learning respectively which may indicate a need to balance this. Future work will explore how to balance this losses, but it may be possible to put a scheduler to enable the loss after a peak. More details on the ablation study can be found in Appendix \ref{appendix:ablationtests}.
\begin{table}[h]
    \small
    \centering
\begin{tabular}{|l|l|l|}
\hline
Models         & FT-CC  & NB \\ \hline
Distributional & 0.6553        & 0.6974      \\ \hline
Attract-Repel  & 0.6993         &       0.6874      \\ \hline
% AuxGAN   & \multicolumn{1}{r|}{0.5932} & \multicolumn{1}{r|}{0.6814} & \multicolumn{1}{r|}{0.6653} \\ \hline
% RetroGAN & \multicolumn{1}{r|}{0.7796} & \multicolumn{1}{r|}{0.7415} & \multicolumn{1}{r|}{0.7555}  \\ \hline
AuxGAN   & 0.7214  & 0.7335 \\ \hline
RetroGAN & \textbf{0.7595}  & \textbf{0.7735}  \\ \hline
\end{tabular}
    \caption{Light-LS accuracy on \cite{horn2014learning}}
  \label{tab:lightlstest}
\end{table}

\section{Conclusion}

This work presents an improvement on post-specialization work through the use of a CycleGAN-like system called RetroGAN. We show that RetroGAN gives improved performance in both the Full (words which were seen in knowledge/constraints) and the Disjoint (words which were \textbf{not} seen in the constraints) evaluation settings for three benchmarks. It additionally has better performance on a downstream lexical simplification task, further confirming its improved generalization ability.  We conclude that RetroGAN is an improved system for post-specializing embeddings for rare and OOK concepts.\footnote{The system and the data can be accessed at \url{https://github.com/pedrocolon93/retrogan.git}}

\section{Acknowledgements}
This work was made possible thanks to the Media Lab Consortium funding.

\bibliographystyle{acl_natbib}
\bibliography{anthology,acl2021}

\begin{thebibliography}{29}
\expandafter\ifx\csname natexlab\endcsname\relax\def\natexlab#1{#1}\fi

\bibitem[{Bojanowski et~al.(2017)Bojanowski, Grave, Joulin, and
  Mikolov}]{bojanowski2017enriching}
Piotr Bojanowski, Edouard Grave, Armand Joulin, and Tomas Mikolov. 2017.
\newblock Enriching word vectors with subword information.
\newblock \emph{Transactions of the Association for Computational Linguistics},
  5:135--146.

\bibitem[{Devlin et~al.(2019)Devlin, Chang, Lee, and
  Toutanova}]{devlin2018bert}
Jacob Devlin, Ming-Wei Chang, Kenton Lee, and Kristina Toutanova. 2019.
\newblock \href {https://doi.org/10.18653/v1/N19-1423} {{BERT}: Pre-training of
  deep bidirectional transformers for language understanding}.
\newblock In \emph{Proceedings of the 2019 Conference of the North {A}merican
  Chapter of the Association for Computational Linguistics: Human Language
  Technologies, Volume 1 (Long and Short Papers)}, pages 4171--4186,
  Minneapolis, Minnesota. Association for Computational Linguistics.

\bibitem[{Faruqui et~al.(2015)Faruqui, Dodge, Jauhar, Dyer, Hovy, and
  Smith}]{faruqui2014retrofitting}
Manaal Faruqui, Jesse Dodge, Sujay~Kumar Jauhar, Chris Dyer, Eduard Hovy, and
  Noah~A. Smith. 2015.
\newblock \href {https://doi.org/10.3115/v1/N15-1184} {Retrofitting word
  vectors to semantic lexicons}.
\newblock In \emph{Proceedings of the 2015 Conference of the North {A}merican
  Chapter of the Association for Computational Linguistics: Human Language
  Technologies}, pages 1606--1615, Denver, Colorado. Association for
  Computational Linguistics.

\bibitem[{Fukuda(2020)}]{NobukazuFukuda2020Based}
Nobukazu Fukuda. 2020.
\newblock Calculation of distributed representation of unknown words based on
  surface similarity with known words
  \url{https://repository.dl.itc.u-tokyo.ac.jp/?action=repository_action_common_download&item_id=54229&item_no=1&attribute_id=14&file_no=1}.

\bibitem[{Gerz et~al.(2016)Gerz, Vuli{\'c}, Hill, Reichart, and
  Korhonen}]{gerz2016simverb}
Daniela Gerz, Ivan Vuli{\'c}, Felix Hill, Roi Reichart, and Anna Korhonen.
  2016.
\newblock \href {https://doi.org/10.18653/v1/D16-1235} {{S}im{V}erb-3500: A
  large-scale evaluation set of verb similarity}.
\newblock In \emph{Proceedings of the 2016 Conference on Empirical Methods in
  Natural Language Processing}, pages 2173--2182, Austin, Texas. Association
  for Computational Linguistics.

\bibitem[{Glava{\v{s}} and {\v{S}}tajner(2015)}]{glavavs2015simplifying}
Goran Glava{\v{s}} and Sanja {\v{S}}tajner. 2015.
\newblock Simplifying lexical simplification: Do we need simplified corpora?
\newblock In \emph{Proceedings of the 53rd Annual Meeting of the Association
  for Computational Linguistics and the 7th International Joint Conference on
  Natural Language Processing (Volume 2: Short Papers)}, pages 63--68.

\bibitem[{Glava{\v{s}} and Vuli{\'c}(2018)}]{glavavs2018explicit}
Goran Glava{\v{s}} and Ivan Vuli{\'c}. 2018.
\newblock Explicit retrofitting of distributional word vectors.
\newblock In \emph{Proceedings of the 56th Annual Meeting of the Association
  for Computational Linguistics (Volume 1: Long Papers)}, pages 34--45.

\bibitem[{Goodfellow et~al.(2014)Goodfellow, Pouget-Abadie, Mirza, Xu,
  Warde-Farley, Ozair, Courville, and Bengio}]{goodfellow2014generative}
Ian Goodfellow, Jean Pouget-Abadie, Mehdi Mirza, Bing Xu, David Warde-Farley,
  Sherjil Ozair, Aaron Courville, and Yoshua Bengio. 2014.
\newblock Generative adversarial nets.
\newblock In \emph{Advances in neural information processing systems}, pages
  2672--2680.

\bibitem[{Hill et~al.(2015)Hill, Reichart, and Korhonen}]{hill2015simlex}
Felix Hill, Roi Reichart, and Anna Korhonen. 2015.
\newblock Simlex-999: Evaluating semantic models with (genuine) similarity
  estimation.
\newblock \emph{Computational Linguistics}, 41(4):665--695.

\bibitem[{Horn et~al.(2014)Horn, Manduca, and Kauchak}]{horn2014learning}
Colby Horn, Cathryn Manduca, and David Kauchak. 2014.
\newblock Learning a lexical simplifier using wikipedia.
\newblock In \emph{Proceedings of the 52nd Annual Meeting of the Association
  for Computational Linguistics (Volume 2: Short Papers)}, pages 458--463.

\bibitem[{Kamath et~al.(2019)Kamath, Pfeiffer, Ponti, Glava{\v{s}}, and
  Vuli{\'c}}]{kamath2019specializing}
Aishwarya Kamath, Jonas Pfeiffer, Edoardo~Maria Ponti, Goran Glava{\v{s}}, and
  Ivan Vuli{\'c}. 2019.
\newblock \href {https://doi.org/10.18653/v1/W19-4310} {Specializing
  distributional vectors of all words for lexical entailment}.
\newblock In \emph{Proceedings of the 4th Workshop on Representation Learning
  for NLP (RepL4NLP-2019)}, pages 72--83, Florence, Italy. Association for
  Computational Linguistics.

\bibitem[{Kim et~al.(2017)Kim, Cha, Kim, Lee, and Kim}]{kim2017learning}
Taeksoo Kim, Moonsu Cha, Hyunsoo Kim, Jung~Kwon Lee, and Jiwon Kim. 2017.
\newblock Learning to discover cross-domain relations with generative
  adversarial networks.
\newblock In \emph{International Conference on Machine Learning}, pages
  1857--1865. PMLR.

\bibitem[{Kingma and Ba(2015)}]{kingma2014adam}
Diederik~P. Kingma and Jimmy Ba. 2015.
\newblock \href {http://arxiv.org/abs/1412.6980} {Adam: {A} method for
  stochastic optimization}.
\newblock In \emph{3rd International Conference on Learning Representations,
  {ICLR} 2015, San Diego, CA, USA, May 7-9, 2015, Conference Track
  Proceedings}.

\bibitem[{Lauscher et~al.(2019)Lauscher, Vuli{\'c}, Ponti, Korhonen, and
  Glava{\v{s}}}]{lauscher2019informing}
Anne Lauscher, Ivan Vuli{\'c}, Edoardo~Maria Ponti, Anna Korhonen, and Goran
  Glava{\v{s}}. 2019.
\newblock Informing unsupervised pretraining with external linguistic
  knowledge.
\newblock \emph{arXiv preprint arXiv:1909.02339}.

\bibitem[{Li et~al.(2020)Li, Jamieson, Rostamizadeh, Gonina, Ben-tzur, Hardt,
  Recht, and Talwalkar}]{MLSYS2020_f4b9ec30}
Liam Li, Kevin Jamieson, Afshin Rostamizadeh, Ekaterina Gonina, Jonathan
  Ben-tzur, Moritz Hardt, Benjamin Recht, and Ameet Talwalkar. 2020.
\newblock \href
  {https://proceedings.mlsys.org/paper/2020/file/f4b9ec30ad9f68f89b29639786cb62ef-Paper.pdf}
  {A system for massively parallel hyperparameter tuning}.
\newblock In \emph{Proceedings of Machine Learning and Systems}, volume~2,
  pages 230--246.

\bibitem[{Mrk{\v{s}}i{\'c} et~al.(2017)Mrk{\v{s}}i{\'c}, Vuli{\'c},
  {\'O}~S{\'e}aghdha, Leviant, Reichart, Ga{\v{s}}i{\'c}, Korhonen, and
  Young}]{mrkvsic2017semantic}
Nikola Mrk{\v{s}}i{\'c}, Ivan Vuli{\'c}, Diarmuid {\'O}~S{\'e}aghdha, Ira
  Leviant, Roi Reichart, Milica Ga{\v{s}}i{\'c}, Anna Korhonen, and Steve
  Young. 2017.
\newblock Semantic specialization of distributional word vector spaces using
  monolingual and cross-lingual constraints.
\newblock \emph{Transactions of the association for Computational Linguistics},
  5:309--324.

\bibitem[{Nair and Hinton(2010)}]{nair2010rectified}
Vinod Nair and Geoffrey~E Hinton. 2010.
\newblock Rectified linear units improve restricted boltzmann machines.
\newblock In \emph{Proceedings of the 27th international conference on machine
  learning (ICML-10)}, pages 807--814.

\bibitem[{Peters et~al.(2019)Peters, Neumann, Logan, Schwartz, Joshi, Singh,
  and Smith}]{peters2019knowledge}
Matthew~E. Peters, Mark Neumann, Robert Logan, Roy Schwartz, Vidur Joshi,
  Sameer Singh, and Noah~A. Smith. 2019.
\newblock \href {https://doi.org/10.18653/v1/D19-1005} {Knowledge enhanced
  contextual word representations}.
\newblock In \emph{Proceedings of the 2019 Conference on Empirical Methods in
  Natural Language Processing and the 9th International Joint Conference on
  Natural Language Processing (EMNLP-IJCNLP)}, pages 43--54, Hong Kong, China.
  Association for Computational Linguistics.

\bibitem[{Pilehvar et~al.(2018)Pilehvar, Kartsaklis, Prokhorov, and
  Collier}]{pilehvar2018card}
Mohammad~Taher Pilehvar, Dimitri Kartsaklis, Victor Prokhorov, and Nigel
  Collier. 2018.
\newblock \href {https://doi.org/10.18653/v1/D18-1169} {Card-660: {C}ambridge
  rare word dataset - a reliable benchmark for infrequent word representation
  models}.
\newblock In \emph{Proceedings of the 2018 Conference on Empirical Methods in
  Natural Language Processing}, pages 1391--1401, Brussels, Belgium.
  Association for Computational Linguistics.

\bibitem[{Ponti et~al.(2018)Ponti, Vuli{\'c}, Glava{\v{s}}, Mrk{\v{s}}i{\'c},
  and Korhonen}]{ponti2018adversarial}
Edoardo~Maria Ponti, Ivan Vuli{\'c}, Goran Glava{\v{s}}, Nikola
  Mrk{\v{s}}i{\'c}, and Anna Korhonen. 2018.
\newblock \href {https://doi.org/10.18653/v1/D18-1026} {Adversarial propagation
  and zero-shot cross-lingual transfer of word vector specialization}.
\newblock In \emph{Proceedings of the 2018 Conference on Empirical Methods in
  Natural Language Processing}, pages 282--293, Brussels, Belgium. Association
  for Computational Linguistics.

\bibitem[{Speer and Chin(2016)}]{speer2016ensemble}
Robyn Speer and Joshua Chin. 2016.
\newblock An ensemble method to produce high-quality word embeddings.
\newblock \emph{arXiv preprint arXiv:1604.01692}.

\bibitem[{Speer et~al.(2017)Speer, Chin, and Havasi}]{speer2017conceptnet}
Robyn Speer, Joshua Chin, and Catherine Havasi. 2017.
\newblock \href {http://aaai.org/ocs/index.php/AAAI/AAAI17/paper/view/14972}
  {{ConceptNet} 5.5: An open multilingual graph of general knowledge}.
\newblock pages 4444--4451.

\bibitem[{Tripathy et~al.(2018)Tripathy, Kannala, and
  Rahtu}]{tripathy2018learning}
Soumya Tripathy, Juho Kannala, and Esa Rahtu. 2018.
\newblock Learning image-to-image translation using paired and unpaired
  training samples.
\newblock In \emph{Asian Conference on Computer Vision}, pages 51--66.
  Springer.

\bibitem[{Vuli{\'c} et~al.(2018)Vuli{\'c}, Glava{\v{s}}, Mrk{\v{s}}i{\'c}, and
  Korhonen}]{vulic2018post}
Ivan Vuli{\'c}, Goran Glava{\v{s}}, Nikola Mrk{\v{s}}i{\'c}, and Anna Korhonen.
  2018.
\newblock \href {https://doi.org/10.18653/v1/N18-1048} {Post-specialisation:
  Retrofitting vectors of words unseen in lexical resources}.
\newblock In \emph{Proceedings of the 2018 Conference of the North {A}merican
  Chapter of the Association for Computational Linguistics: Human Language
  Technologies, Volume 1 (Long Papers)}, pages 516--527, New Orleans,
  Louisiana. Association for Computational Linguistics.

\bibitem[{Vuli{\'c} and Mrk{\v{s}}i{\'c}(2018)}]{vulic2017specialising}
Ivan Vuli{\'c} and Nikola Mrk{\v{s}}i{\'c}. 2018.
\newblock \href {https://doi.org/10.18653/v1/N18-1103} {Specialising word
  vectors for lexical entailment}.
\newblock In \emph{Proceedings of the 2018 Conference of the North {A}merican
  Chapter of the Association for Computational Linguistics: Human Language
  Technologies, Volume 1 (Long Papers)}, pages 1134--1145, New Orleans,
  Louisiana. Association for Computational Linguistics.

\bibitem[{Weston et~al.(2011)Weston, Bengio, and Usunier}]{weston2011wsabie}
Jason Weston, Samy Bengio, and Nicolas Usunier. 2011.
\newblock Wsabie: Scaling up to large vocabulary image annotation.
\newblock In \emph{Twenty-Second International Joint Conference on Artificial
  Intelligence}.

\bibitem[{Yang et~al.(2019)Yang, Zhu, Sachidananda, and Darve}]{yang2019out}
Ziyi Yang, Chenguang Zhu, Vin Sachidananda, and Eric Darve. 2019.
\newblock Out-of-vocabulary embedding imputation with grounded language
  information by graph convolutional networks.
\newblock \emph{arXiv e-prints}, pages arXiv--1906.

\bibitem[{Ye et~al.(2019)Ye, Chen, Wang, and Ling}]{ye2019align}
Zhi-Xiu Ye, Qian Chen, Wen Wang, and Zhen-Hua Ling. 2019.
\newblock Align, mask and select: A simple method for incorporating commonsense
  knowledge into language representation models.
\newblock \emph{arXiv preprint arXiv:1908.06725}.

\bibitem[{Zhu et~al.(2017)Zhu, Park, Isola, and Efros}]{zhu2017unpaired}
Jun-Yan Zhu, Taesung Park, Phillip Isola, and Alexei~A Efros. 2017.
\newblock Unpaired image-to-image translation using cycle-consistent
  adversarial networks.
\newblock In \emph{Proceedings of the IEEE international conference on computer
  vision}, pages 2223--2232.

\end{thebibliography}
\clearpage
\appendix
\section{Ablation Tests}
\begin{figure*}[ht!]
     \begin{center}
% The full tests
        \subfigure[SimLex Ablation Test-Full Scenario]{%
            \label{fig:sf1}
            \includegraphics[width=0.3\textwidth]{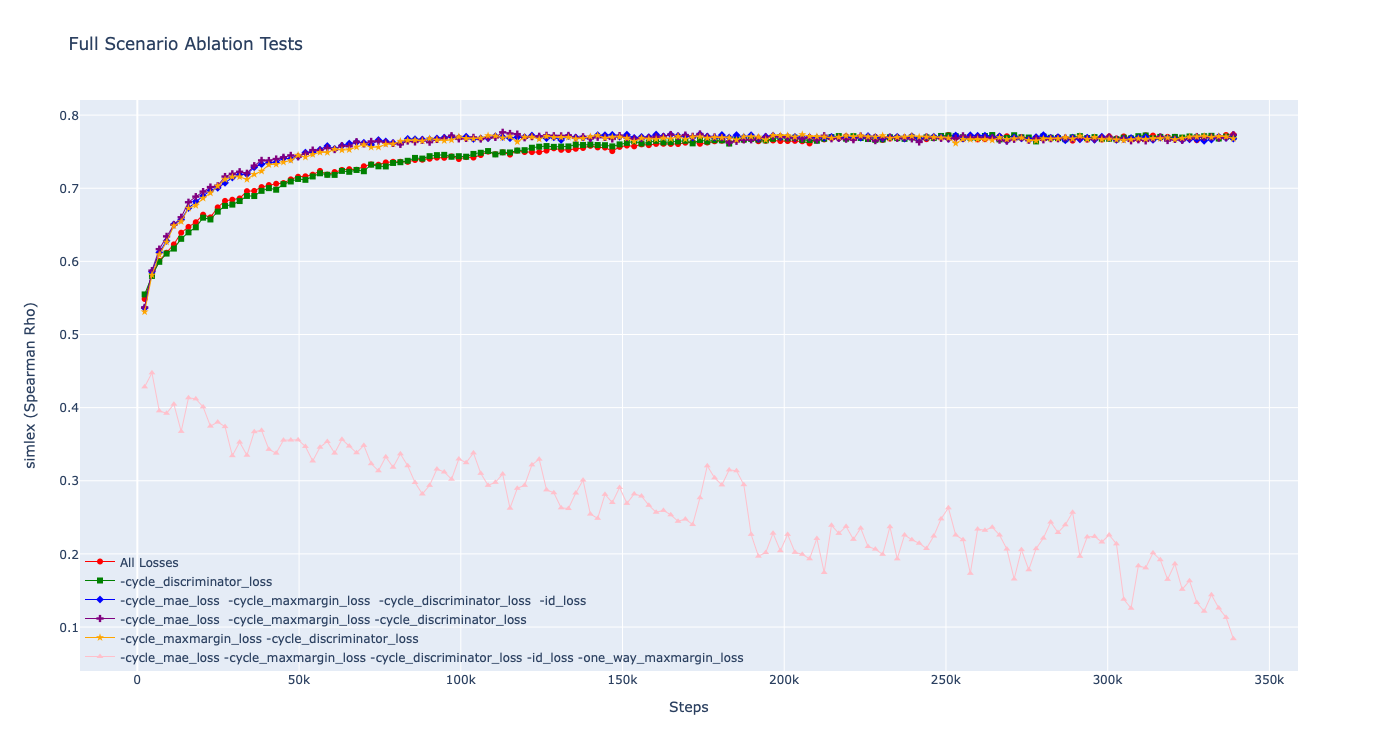}
        }%
        \subfigure[SimVerb Ablation Test-Full Scenario]{%
            \label{fig:sf2}
            \includegraphics[width=0.3\textwidth]{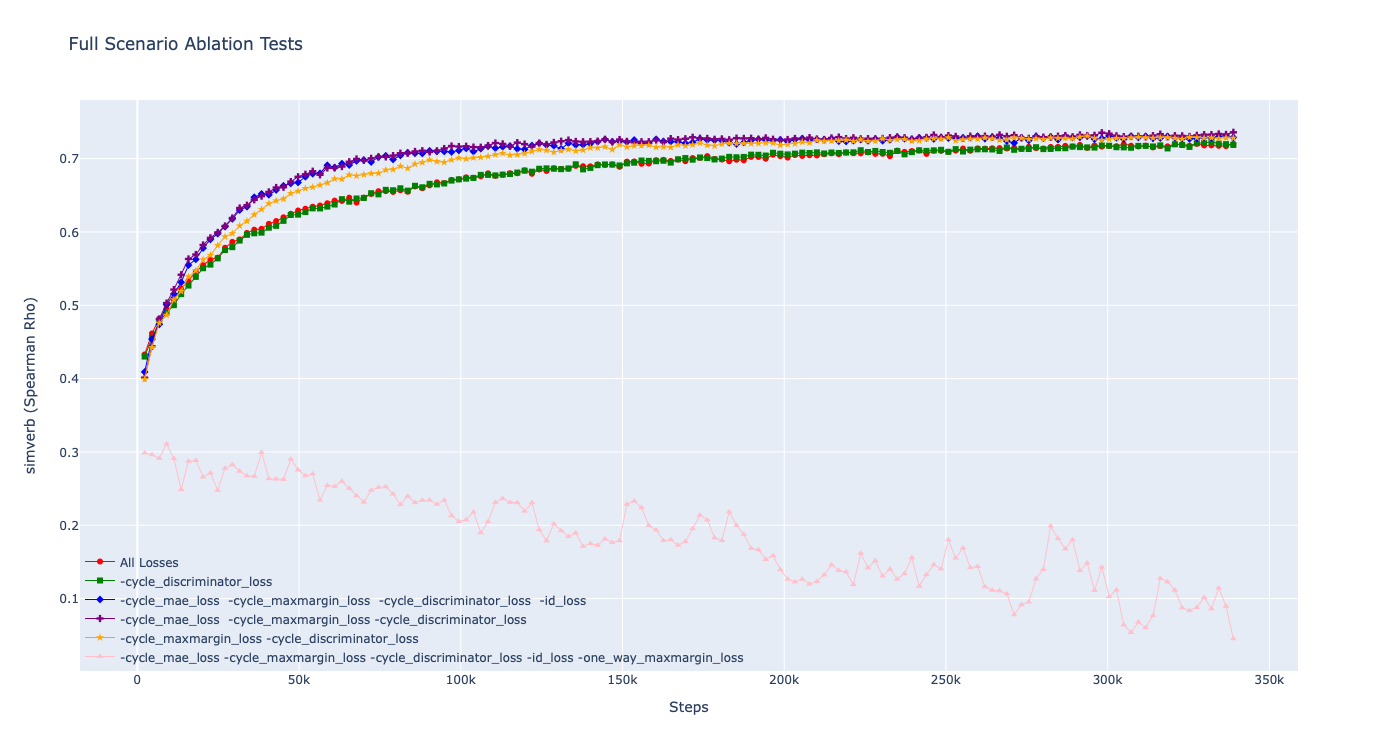}
        }%
        \subfigure[CARD-660 Ablation Test-Full Scenario]{%
            \label{fig:sf3}
            \includegraphics[width=0.3\textwidth]{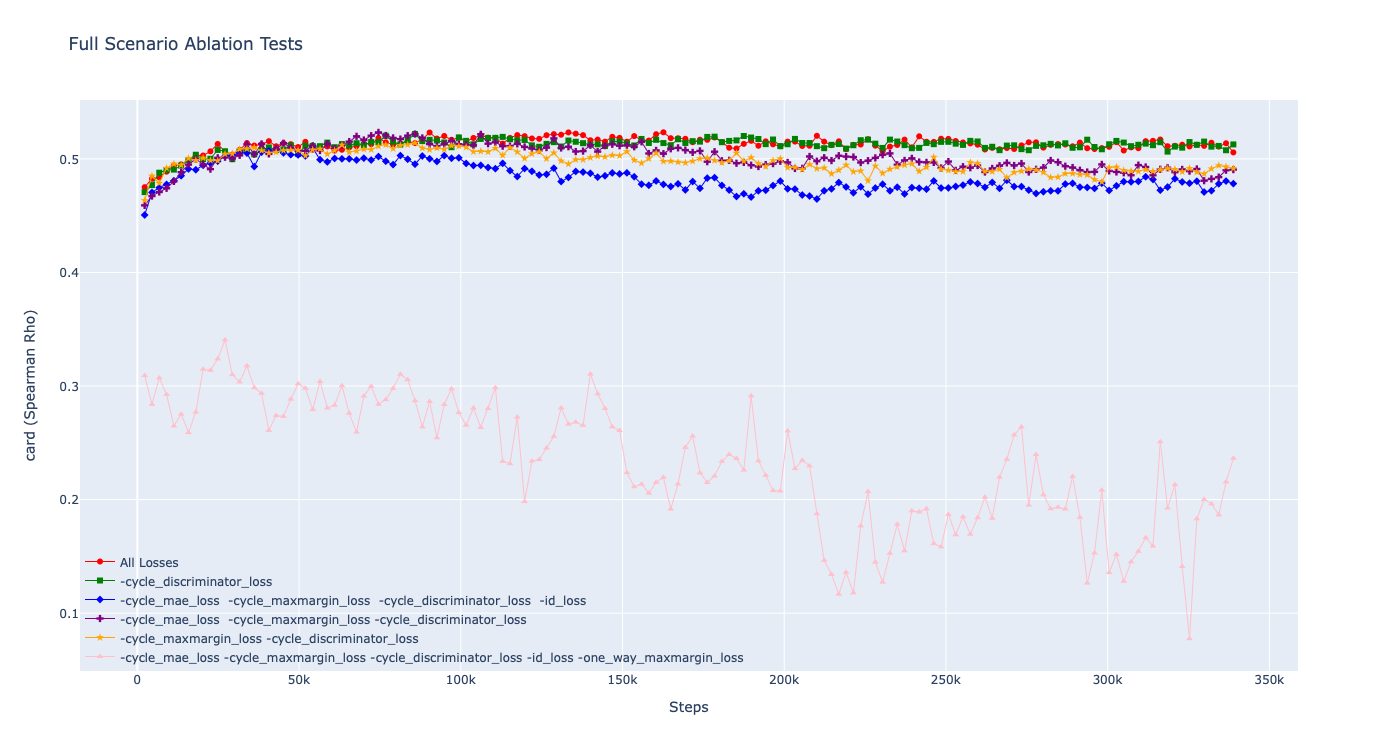}
        }\\ %  ------- End of the first row 
% The disjoint tests
        \subfigure[SimLex Ablation Test-Disjoint Scenario]{%
           \label{fig:sf4}
           \includegraphics[width=0.3\textwidth]{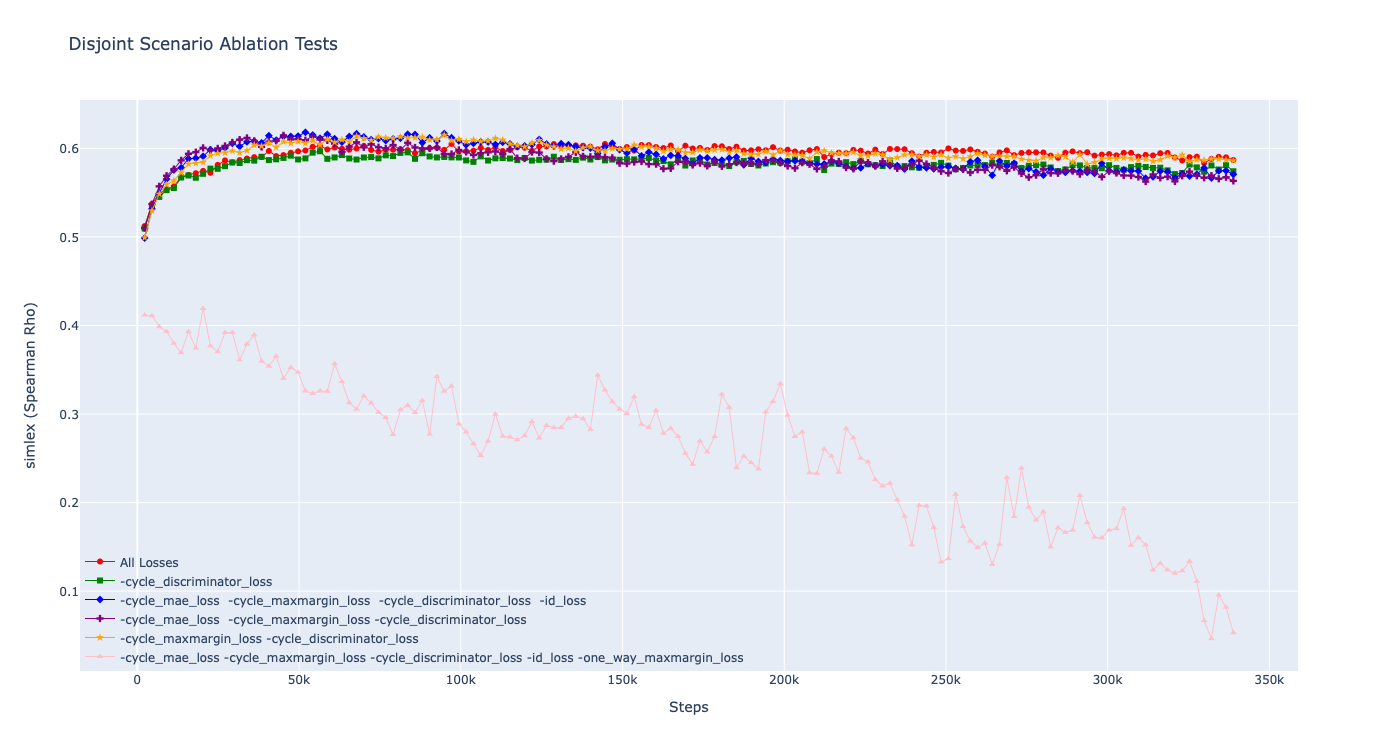}
        }

        \subfigure[SimVerb Ablation Test-Disjoint Scenario]{%
            \label{fig:sf5}
            \includegraphics[width=0.3\textwidth]{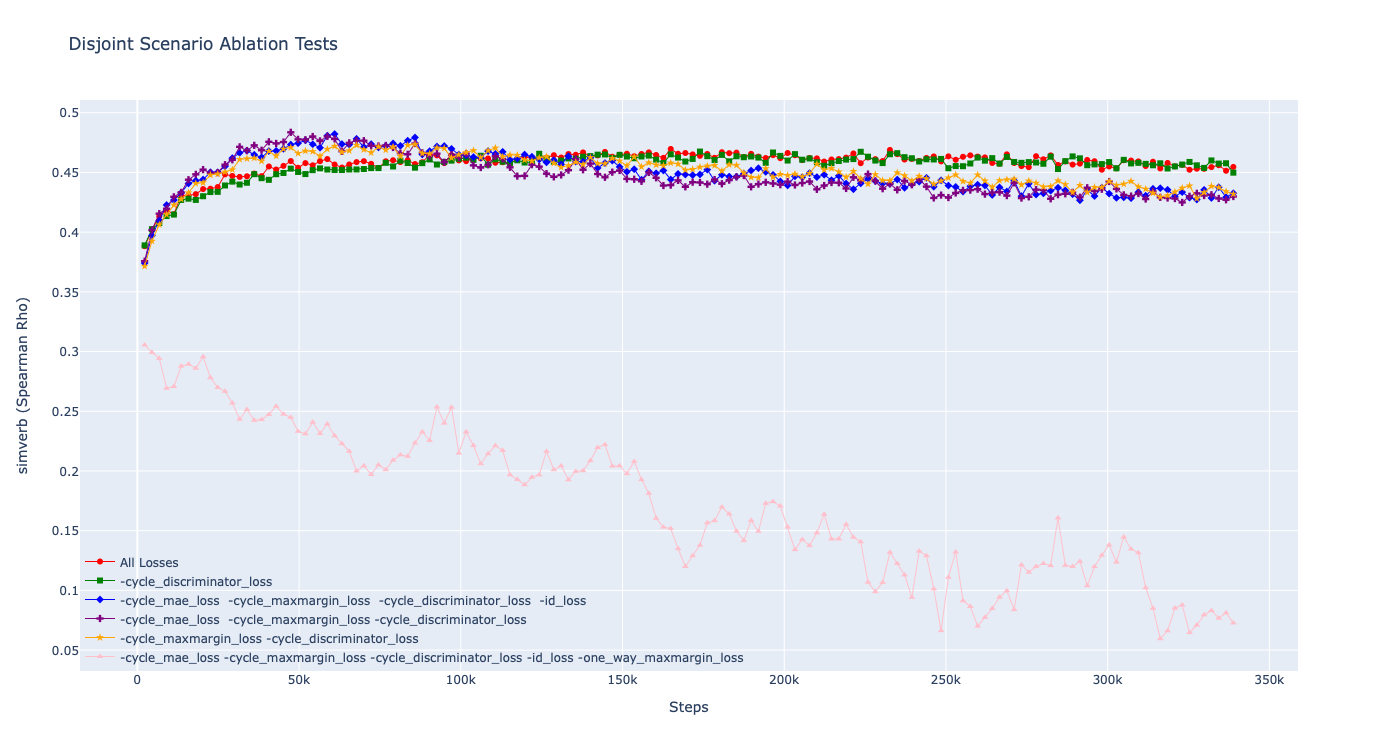}
        }%
        \subfigure[CARD-660 Ablation Test-Disjoint Scenario]{%
           \label{fig:sf6}
           \includegraphics[width=0.3\textwidth]{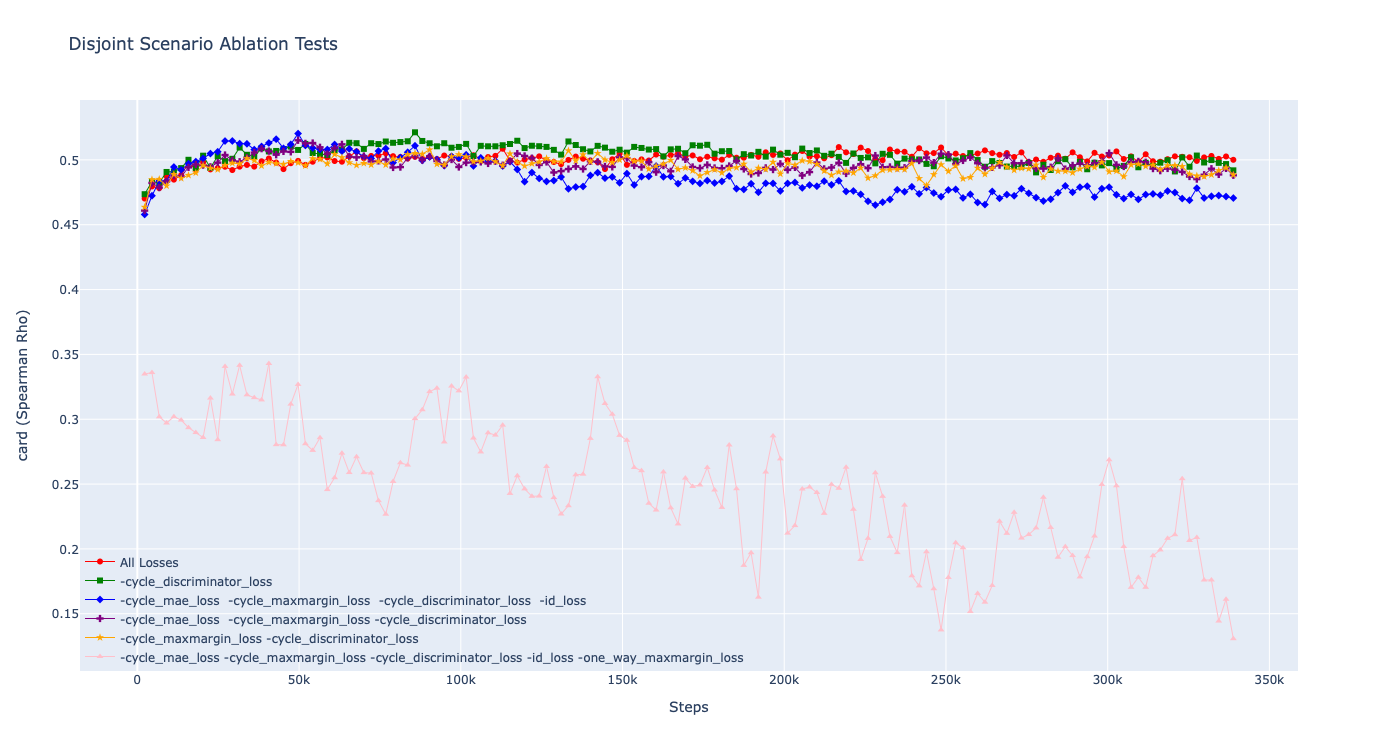}
        }\\
% Toggle full tests
        \subfigure[SimLex Ablation Toggle Test-Disjoint Scenario]{%
           \label{fig:sf4}
           \includegraphics[width=0.3\textwidth]{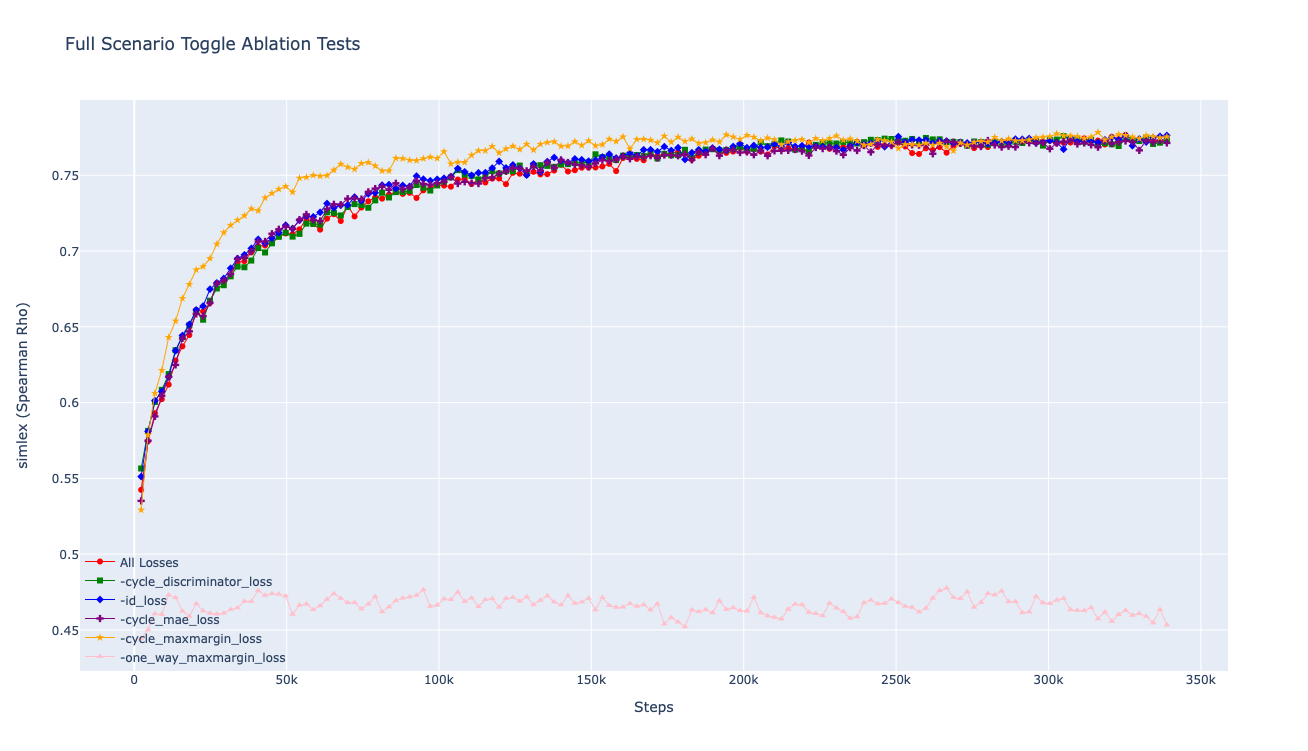}
        }
        \subfigure[SimVerb Ablation Toggle Test-Disjoint Scenario]{%
            \label{fig:sf5}
            \includegraphics[width=0.3\textwidth]{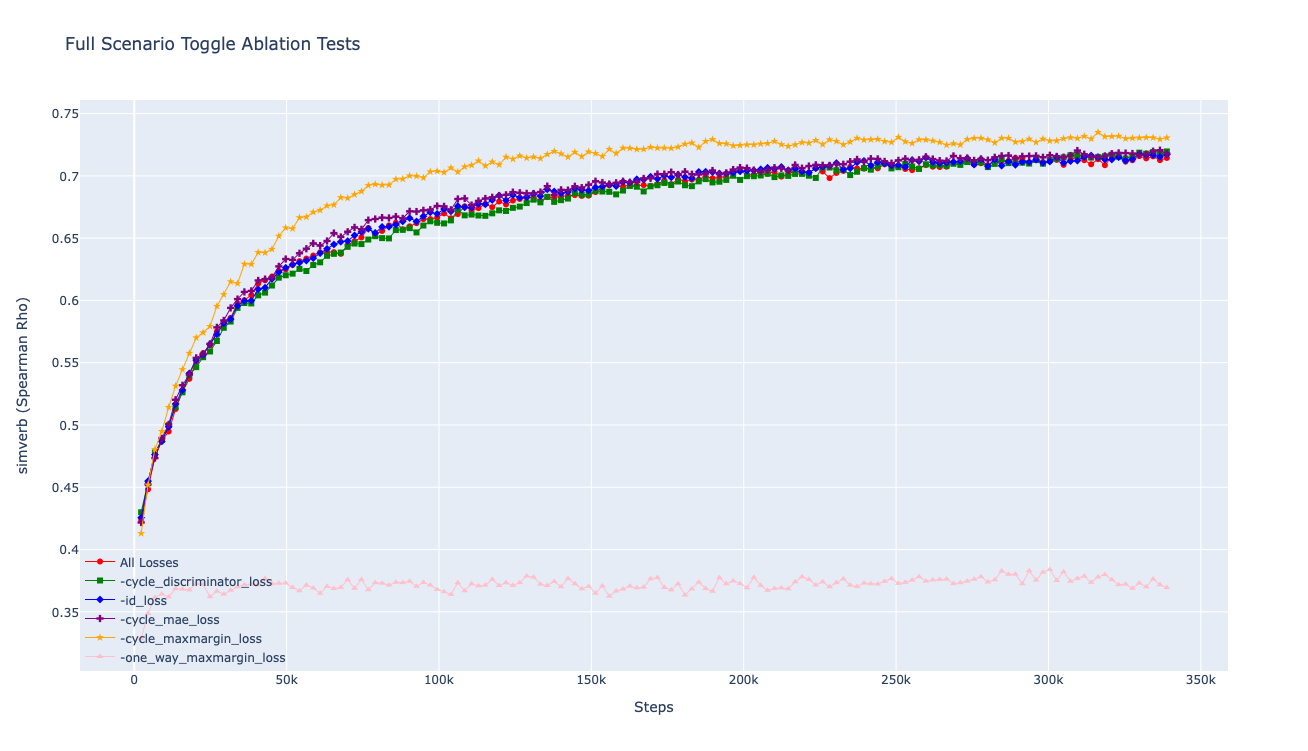}
        }%
        \subfigure[CARD-660 Ablation Toggle Test-Disjoint Scenario]{%
           \label{fig:sf6}
           \includegraphics[width=0.3\textwidth]{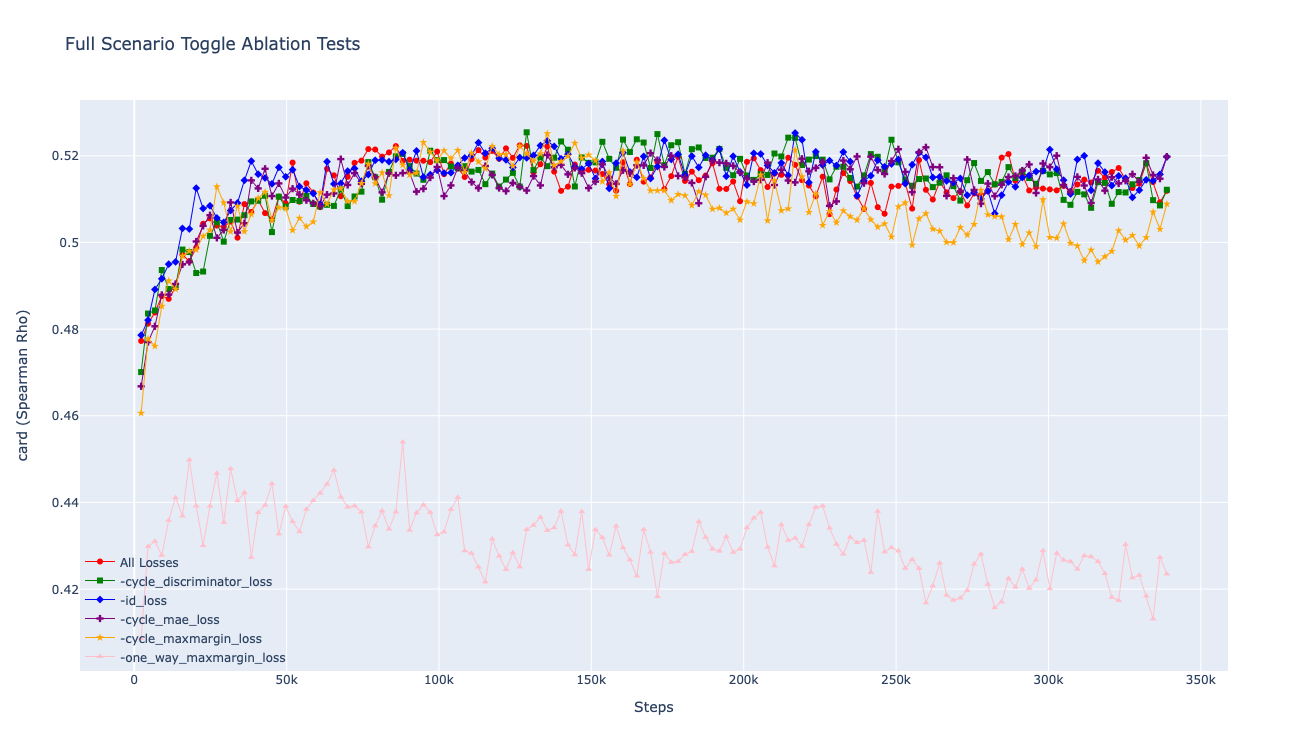}
        }\\
         %  ------- End of the first row 
%Toggle disjoint
        \subfigure[SimLex Ablation Toggle Test-Disjoint Scenario]{%
           \label{fig:sf4}
           \includegraphics[width=0.3\textwidth]{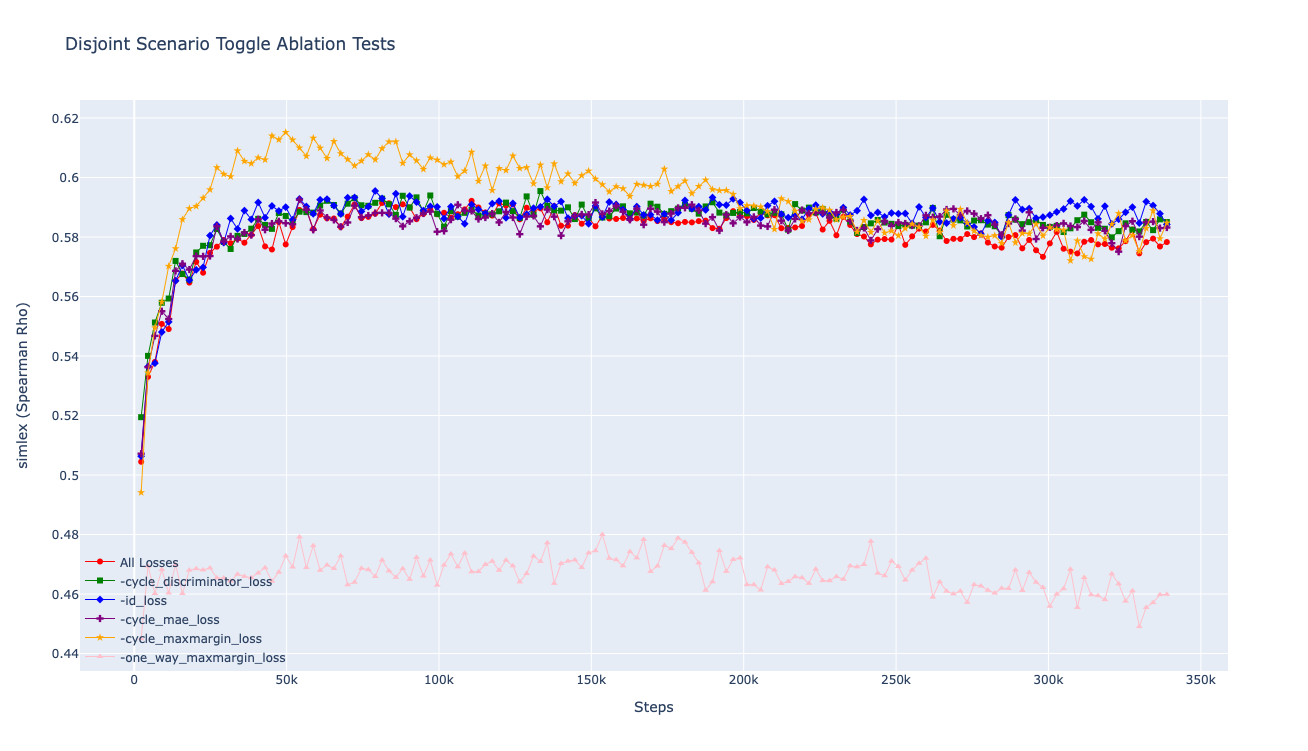}
        }
        \subfigure[SimVerb Ablation Toggle Test-Disjoint Scenario]{%
            \label{fig:sf5}
            \includegraphics[width=0.3\textwidth]{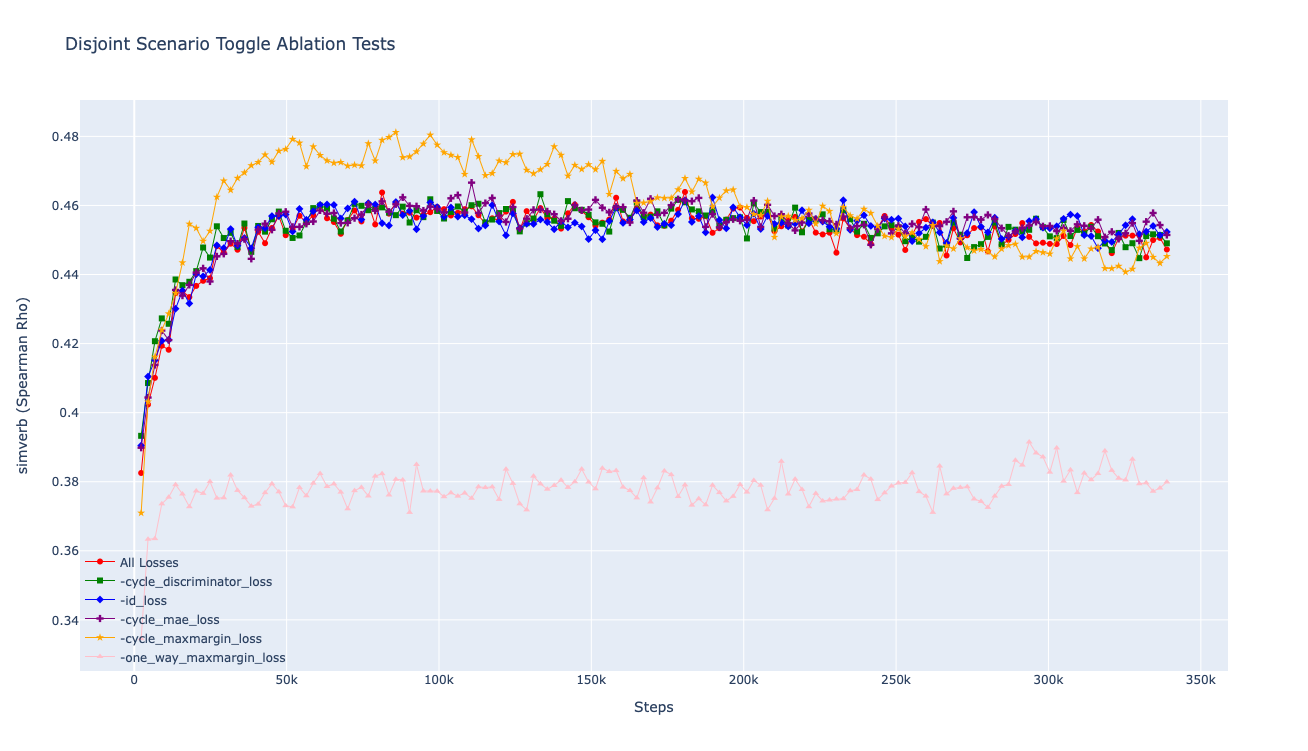}
        }%
        \subfigure[CARD-660 Ablation Toggle Test-Disjoint Scenario]{%
           \label{fig:sf6}
           \includegraphics[width=0.3\textwidth]{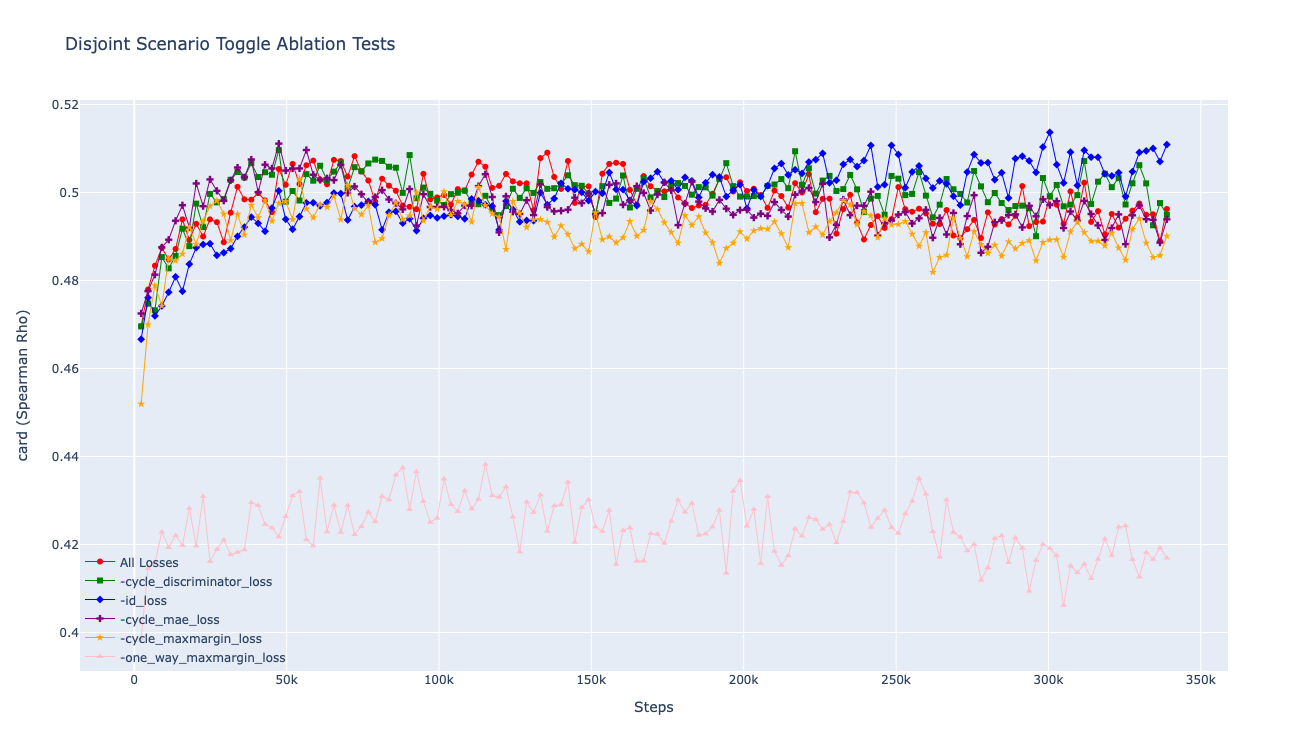}
        }
    \end{center}
    \caption{%
        Ablation test results for SimLex, SimVerb, and CARD-660.  A higher resolution version can be found in the repository at: \url{https://github.com/pedrocolon93/retrogan.git}
     }
   \label{fig:subfigures}
\end{figure*}

We performed a small ablation study to examine how the multiple losses in RetroGAN affect performance.  A one-by-one removal of these can be seen in figures \ref{fig:sf1},\ref{fig:sf2},\ref{fig:sf3},\ref{fig:sf4},\ref{fig:sf5},\ref{fig:sf6}. A toggle of each of the losses can be seen in: \ref{fig:sf1},\ref{fig:sf2},\ref{fig:sf3},\ref{fig:sf4},\ref{fig:sf5},\ref{fig:sf6}.  The difference between the toggle and one-by-one removal is that in the toggle, we simple turn off the specified loss and leave the others untouched, whereas in the one-by-one removal we turn off one-by-one the losses, in this way we can see the individual effects, and the group effects. We evaluated the FT-CC and the Attract-Repel retrofitted FT-CC in the same scenarios as the evaluations before (Disjoint and Full).  We note that the Disjoint setting for Card includes some of the words in the constraints. 

The max margin loss utilized by \cite{ponti2018adversarial} (\textit{one\_way\_maxmargin\_loss}) is essential for high performance on the datasets.  Without this loss, in all of the figures, we see that the scores in all our tests fall by at least by 0.1.  This is seen in both the toggle and the one-by-one case. We can also see that the Cyclic version of this loss (\textit{cycle\_maxmargin\_loss}) slows down learning initially, but stabilizes it in later iterations. We can see that by removing it we get higher performance in earlier iterations but the performance decays as more iterations are given. This may be because it tries to enforce that the semantic components of the embeddings be similar after going through the cycle, but it may be a hard objective to achieve.  This loss is especially useful for the rare-word Card-660 evaluation. By looking at the toggle ablation test for this loss, we can see that it indeed can lead to better earlier performance, however it decays with time.  

The identity loss (\textit{id\_loss}) helps to stabilize the training in later iterations.  Removal of this loss significantly affects the disjoint settings, and the reason for it may be that it gives some indication of the important semantic components of the vectors that are being post-specialized.  This in the disjoint setting leads to significant performance reductions on the later iterations. Interestingly enough, by toggling off only this loss, we can see that it leads to better performance, which means that with all the other losses, it may contain redundant information that may hinder performance, however if the model relies on the loss without other losses, its information is useful. 

The Cycle Conditional discriminator loss (\textit{cycle\_discriminator\_loss}) also contributes to the stability and generalization of the later learning.  Removing this loss does not improve early learning, save on the Card-660 dataset, and in most of the other tests, there is not a large noticeable difference.  However, on the disjoint setting we do see that it performance decays in later iterations. We suspect the conditioning helps slightly in the stabilization, and generalization of the system, but its effect is not too much. 

The Cycle Loss (\textit{cycle\_mae\_loss}), also stabilizes and helps in the generalization of our system.  We can see in the disjoint settings in particular, that its removal hinders the model in later iterations. We suspect that since the consistency is not being enforced, the model does not learn effectively to preserve important, possibly non-semantic, parts from the distributional and the retrofitted domain. 

As a practical recommendation, we suggest removing the cyclic max-margin loss either completely (pausing the training early at it's peak around 50k-100k iterations), or toggling it after this initial training to get the speedup and the generation. Another practical recommendation may be to disable the identity loss all by itself. The other losses can be maintained as they are described in this work.

\label{appendix:ablationtests}
\section{Out-of-knowledge Scalability Tests}
In table \ref{table:fulloovtestsapp} we test the performance of post specialization as more constraints are added into the retrofitting process. We note that AuxGAN's performance saturates after 50\% whereas RetroGAN keeps learning, albeit less accurately than the retrofitting system.  These tests were run for 100k batches on RetroGAN and for 10M iterations (312500 RetroGAN batches) on AuxGAN. 
\label{appendix:fulloovtest}
\section{Additional embedding pre-processing}
Input and output vectors are divided by the Euclidean (2) norm.  This helps slightly in the performance of the semantic comparison benchmarks. No other pre-processing is done on the vectors.     

\section{Architecture Details}

\begin{figure}[hb!]
  \centering
  \includegraphics[width=0.5\textwidth]{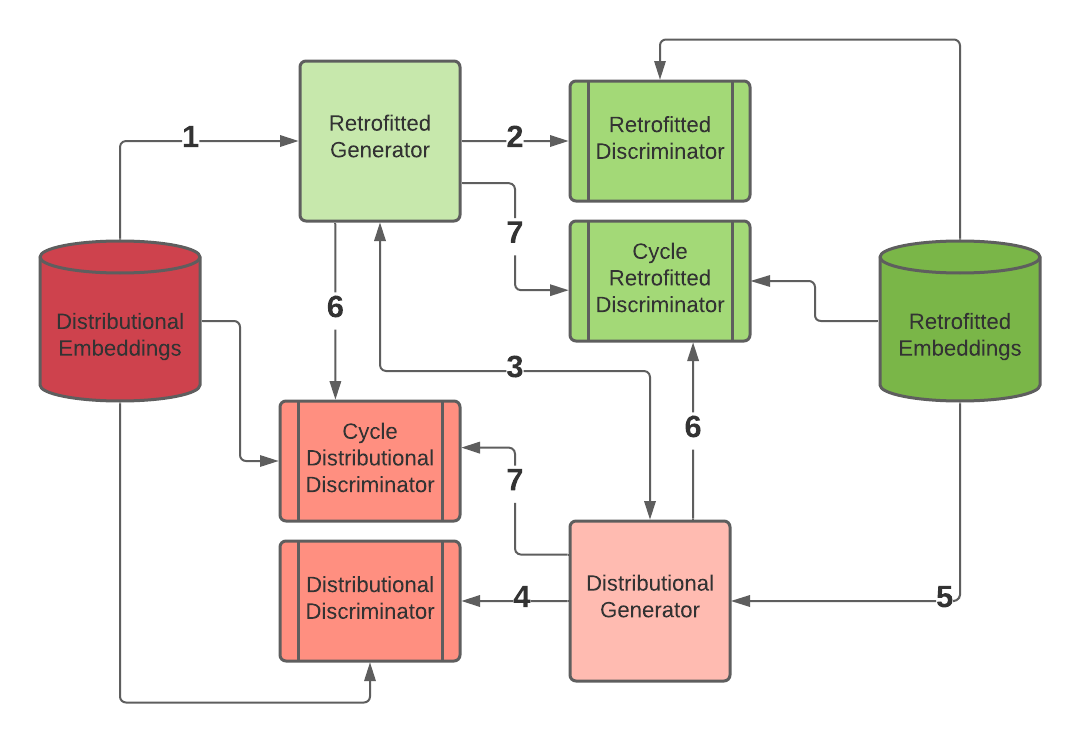}
  \caption{Architecture diagram for RetroGAN}
  \label{fig:architecturerg}
\end{figure}

In figure \ref{fig:architecturerg}, we can see the architecture that RetroGAN uses.  On a training step, the losses are calculated as follows. For the cyclic losses, the system samples embeddings from the distributional embeddings and their retrofitted counterparts and these samples are passed to the generators (1, 5 in the figure).  Then, the generators' output is passed to the counterpart generator (Distributional Generator passes to Retrofitted Generator and vice versa, seen as 3 in the figure).  The output of this is then used to calculate the max margin loss, and passed on to the subsequent discriminator to calculate the cycle discriminator loss (2, 4 in the figure). In addition to this, after going through the cycle of generators (1 or 5, 3 in the figure) we train the conditional discriminators by conditioning on real inputs from the retrofitted or distributional embeddings, or by conditioning on fake inputs (6,7 in the figure).   

The amount of parameters in each model and the layers can be found in table \ref{table:architecturespecs}.

\section{Parameter Tuning}
We performed a parameter tuning using the Ray tuning library, to try and generate a configuration that would be optimal for RetroGAN. We utilized the ASHA Scheduler\cite{MLSYS2020_f4b9ec30}
along with the following search space configuration:
\begin{verbatim}
config = {
"g_lr" : tune.qloguniform(0.00005,
.1,0.00005),
"d_lr" : tune.qloguniform(0.00005,
.1,0.00005),
"one_way_mm":True,
"cycle_mm":True,
"cycle_dis":True,
"id_loss":True,
"cycle_loss":True,
"batch_size":tune.choice([16,32,64]),
"generator_size":tune.choice([512,
1024,2048]),
"discriminator_size":
tune.choice([512,1024,2048]),
"generator_hidden_layers": 
tune.choice([1,2,3]),
"discriminator_hidden_layers": 
tune.choice([1,2,3]),
"dis_train_amount":
tune.choice([1,2,3])
}
\end{verbatim}
We used the SimVerb score to guide the parameter optimization, because it was the score that involved the largest sample of words.  We also utilized a machine with two Intel processors with 48 cores in total, 128GB of RAM, an NVIDIA P6000 and a NVIDIA 1080TI. We used 25 samples in the optimization due to time constraints, although this value can be expanded more. We also ran this for 35 epochs, because the performance after that would become relatively stable and not increase greatly.  

The best results from the optimization are the following:
\begin{verbatim}
{
'g_lr': 0.00495, 
'd_lr': 0.00885,
'one_way_mm': True, 
'cycle_mm': True,
'cycle_dis': True, 
'id_loss': True,
'cycle_loss': True, 
'batch_size': 32,
'generator_size': 2048, 
'discriminator_size': 2048,
'generator_hidden_layers': 1,
'discriminator_hidden_layers': 3,
'dis_train_amount': 1
}
\end{verbatim}

\begin{table*}[]
\begin{tabular}{l|l|l|l|l|l|l|l|l|l|}
\cline{2-10}
                                    & \multicolumn{3}{r|}{5\%}                         & \multicolumn{3}{r|}{10\%}                        & \multicolumn{3}{r|}{25\%}                        \\ \hline
\multicolumn{1}{|l|}{Models}        & SL             & SV             & C660           & SL             & SV             & C660           & SL             & SV             & C660           \\ \hline
\multicolumn{1}{|l|}{Attract-Repel} & 0.347          & 0.355          & 0.113          & 0.550          & 0.589          & 0.187          & 0.701          & 0.700          & 0.217          \\ \hline
\multicolumn{1}{|l|}{AuxGAN}        & 0.615          & 0.510          & 0.453          & 0.667          & 0.569          & 0.470          & 0.679          & 0.581          & 0.475          \\ \hline
\multicolumn{1}{|l|}{RetroGAN}      & \textbf{0.624} & \textbf{0.538} & \textbf{0.489} & \textbf{0.701} & \textbf{0.652} & \textbf{0.493} & \textbf{0.738} & \textbf{0.690} & \textbf{0.502} \\ \hline
\end{tabular}
\begin{tabular}{l|l|l|l|l|l|l|l|l|l|}
\cline{2-10}
                                    & \multicolumn{3}{r|}{50\%}                        & \multicolumn{3}{r|}{75\%}                        & \multicolumn{3}{r|}{100\%}                       \\ \hline
\multicolumn{1}{|l|}{Models}        & SL             & SV             & C660           & SL             & SV             & C660           & SL             & SV             & C660           \\ \hline
\multicolumn{1}{|l|}{Attract-Repel} & \textbf{0.759} & \textbf{0.747} & 0.252          & \textbf{0.766} & \textbf{0.757} & 0.244          & \textbf{0.771} & \textbf{0.761} & 0.257          \\ \hline
\multicolumn{1}{|l|}{AuxGAN}        & 0.685          & 0.600          & 0.490          & 0.688          & 0.597          & 0.480          & 0.690          & 0.601          & 0.486          \\ \hline
\multicolumn{1}{|l|}{RetroGAN}      & 0.755          & 0.716          & \textbf{0.511} & 0.763          & 0.721          & \textbf{0.507} & 0.762          & 0.715          & \textbf{0.509} \\ \hline
\end{tabular}
\caption{Performance of post specialization and retrofitting as more constraints are added to a system.}
\label{table:fulloovtestsapp}
\end{table*}

\begin{table*}[]
\small
\centering
\begin{tabular}{|l|l|l|}
\hline
Model                                                                      & Amount of Parameters & Layers                                                                                                                                                             \\ \hline\hline
Generator                                                                  & 5,427,500(*2)        & \begin{tabular}[c]{@{}l@{}}Linear (300x2048), ReLU, Dropout (0.3),\\ Linear (2048x2048),ReLU \\ Dropout (0.3), Linear (2048x300)\end{tabular}                      \\ \hline
Discriminator                                                              & 4,818,945(*2)        & \begin{tabular}[c]{@{}l@{}}Linear (300x2048), ReLU, Dropout (0.3),\\ Linear (2048x2048),ReLU, Batch Norm, \\ Dropout (0.3), Linear (2048x1), Sigmoid\end{tabular}  \\ \hline
\begin{tabular}[c]{@{}l@{}}Cycle Conditional \\ Discriminator\end{tabular} & 5,433,345(*2)        & \begin{tabular}[c]{@{}l@{}}Linear (600x2048), ReLU, Dropout (0.3),\\ Linear (2048x2048), ReLU, Batch Norm, \\ Dropout (0.3), Linear (2048x1), Sigmoid\end{tabular} \\ \hline\hline
Total                                                                      & 31,359,580           & -                                                                                                                                                                  \\ \hline
\end{tabular}
\caption{Amount of trainable parameters and layers in RetroGAN}
\label{table:architecturespecs}

\end{table*}

\end{document}